\documentclass{article}

\PassOptionsToPackage{numbers, compress}{natbib}
\usepackage[final]{neurips_2021}

\usepackage[utf8]{inputenc} 
\usepackage[T1]{fontenc}    
\usepackage[pagebackref=false,breaklinks=true,colorlinks,bookmarks=false]{hyperref}
\usepackage{url}            
\usepackage{booktabs}       
\usepackage{amsfonts}       
\usepackage{nicefrac}       
\usepackage{microtype}      
\usepackage{xcolor}         
\usepackage{amsmath}
\DeclareMathOperator{\E}{\mathbb{E}}
\newcommand{\norm}[1]{\left\lVert#1\right\rVert}
\newcommand{\bs}[1]{\boldsymbol{#1}}

\usepackage{multirow}
\usepackage{amsthm}
\usepackage{amsmath}
\usepackage{amssymb}
\usepackage{algorithm}
\usepackage{algorithmic}

\usepackage{amsmath,amssymb}
\usepackage{array}
\usepackage{booktabs} 
\usepackage{epsfig}
\usepackage{graphicx}
\usepackage{bmpsize}
\usepackage{makecell}
\usepackage{pifont}
\newcommand{\cmark}{\ding{51}}%
\newcommand{\xmark}{\ding{55}}%
\usepackage{pifont}

\newcommand{\s}{\boldsymbol{s}}

\newcommand{\x}{\boldsymbol{x}}

\newcommand{\calC}{\mathcal{C}} 
\newcommand{\calF}{\mathcal{F}}

\newcommand{\rE}{\mathbb{E}}

\allowdisplaybreaks

\usepackage{listings}
\usepackage{xcolor}

\definecolor{codegreen}{rgb}{0,0.6,0}
\definecolor{codegray}{rgb}{0.5,0.5,0.5}
\definecolor{codepurple}{rgb}{0.58,0,0.82}
\definecolor{backcolour}{rgb}{0.95,0.95,0.92}

\lstdefinestyle{mystyle}{
    backgroundcolor=\color{backcolour},   
    commentstyle=\color{codegreen},
    keywordstyle=\color{magenta},
    numberstyle=\tiny\color{codegray},
    stringstyle=\color{codepurple},
    basicstyle=\ttfamily\footnotesize,
    breakatwhitespace=false,         
    breaklines=true,                 
    captionpos=b,                    
    keepspaces=true,                 
    numbers=left,                    
    numbersep=5pt,                  
    showspaces=false,                
    showstringspaces=false,
    showtabs=false,                  
    tabsize=2
}

\usepackage[numbers]{natbib}

\title{Efficient Neural Network Training via \\ Forward and Backward Propagation Sparsification}

\author{%
  Xiao Zhou\thanks{Equal contribution}  $~^{1}$, Weizhong Zhang$^{*1}$, Zonghao Chen$^2$, Shizhe Diao$^1$, Tong Zhang\thanks{Jointly with Google Research}$~^1$ \\
  $^1$ Hong Kong University of Science and Technology, $^2$ Tsinghua University\\
  \texttt{xzhoubi@connect.ust.hk, zhangweizhongzju@gmail.com}\\
  \texttt{czh17@mails.tsinghua.edu.cn, sdiaoaa@ust.hk, tongzhang@tongzhang-ml.org} \\
}

\newtheorem{theorem}{Theorem}
\newtheorem{remark}{Remark}
\newtheorem{property}{Property}
\begin{document}

\maketitle

\begin{abstract}
  Sparse training is a natural idea to accelerate the training speed of deep neural networks and save the memory usage, especially since large modern neural networks are significantly over-parameterized.  However, most of the existing methods cannot achieve this goal in practice because 
   the chain rule based gradient (w.r.t. structure parameters) estimators adopted by previous methods require dense computation at least in the backward propagation step.  This paper solves this problem by proposing an efficient sparse training method with completely sparse forward and backward passes. We first formulate the training process as a continuous minimization problem under global sparsity constraint. We then separate the optimization process into two steps, corresponding to weight update and structure parameter update. For the former step, we use the conventional chain rule, which can be sparse via exploiting the sparse structure.  For the latter step, instead of using the chain rule based gradient estimators as in existing methods, we propose a variance reduced policy gradient estimator, which only requires two forward passes without backward propagation, thus achieving completely sparse training. We prove that the variance of our gradient estimator is bounded. Extensive experimental results on real-world datasets demonstrate that compared to previous methods, our algorithm is much more effective in accelerating the training process, up to an order of magnitude faster. 
   

\end{abstract}

\section{Introduction}\label{sec:intro}
  In the last decade, deep neural networks (DNNs) \cite{simonyan2014very,he2016deep,NIPS2017_3f5ee243} have proved their outstanding performance in various fields such as computer vision and natural language processing. However, training such large-sized networks is still very challenging, requiring huge computational power and storage. This hinders us from exploring larger networks, which are likely to have better performance. Moreover, it is a widely-recognized property that modern neural networks  are  significantly overparameterized, which means that a fully trained network can always be sparsified dramatically by network pruning techniques  \cite{han2015deep,guo2016dynamic,liu2018rethinking,zeng2018mlprune,li2016pruning}  into a small sub-network with negligible degradation in accuracy. After pruning, the inference efficiency can be greatly improved. Therefore, a natural question is {\it can we exploit this sparsity to improve the training efficiency?}
  
The emerging technique called sparse network training \cite{han2015learning} is closely related with our question, which can obtain sparse networks  by training from scratch. We can divide existing methods into two categories, i.e., {\it parametric} and {\it non-parametric}, based on whether they explicitly parameterize network structures with trainable variables (termed {\it structure parameters}). Empirical results \cite{liu2017learning,savarese2019winning, yuan2020growing,liu2021sparse} demonstrate that the sparse networks they obtain have comparable accuracy with those obtained from network pruning. 
However, most of them narrowly aim at finding a sparse subnetwork instead of simultaneously sparsifying the computation of training by exploiting the sparse structure. As a consequence, it is hard for them to effectively accelerate the training process in practice on general  platforms, e.g., Tensorflow \cite{abadi2016tensorflow} and Pytorch \cite{paszke2019pytorch}. Detailed reasons are  discussed below:
 \begin{itemize}
     \item Non-parametric methods find the sparse network by repeating a two-stage procedure that alternates between weight optimization and pruning \cite{han2015learning,frankle2019stabilizing}, or by adding a proper sparsity-inducing regularizer on the weights to the objective \cite{liu2015sparse,wen2016learning}. The two-stage methods prune the networks in weight space and usually require retraining the obtained subnetwork from scratch every time when new weights are pruned, which makes training process even more time-consuming. Moreover, the computation of regularized methods is dense since the gradients of a zero-valued weights/filters are still nonzero. 
     \item All the parametric approaches estimate the gradients based on chain rule. The gradient w.r.t. the structure parameters can be nonzero even when the corresponding channel/weight is pruned. Thus, to calculate the gradient via backward propagation, the error has to be propagated through all the neurons/channels. This  means that the computation of backward propagation has to be dense. Concrete analysis can be found in Section \ref{sec:discussion}.
 \end{itemize}
 
We notice that some existing methods \cite{dettmers2019sparse,lym2019prunetrain}  can achieve training speedup by careful implementation. For example,  the dense to sparse algorithm \cite{lym2019prunetrain}  removes some channels if the corresponding weights are quite small for a long time. However,  these methods always need to work with a large model at the beginning epochs and consume huge memory and heavy computation in the early stage. Therefore, even with such careful implementations, the speedups they can achieve are still limited.

 
 
 
 In this paper, we propose an efficient channel-level parametric sparse neural network training method, which is comprised of  \textbf{completely sparse} (See Remark \ref{remark-speedup}) forward and backward propagation. We adopt channel-level sparsity since such sparsity can be efficiently implemented on the current training platforms to save the computational cost. In our method,  we first parameterize the network structure by associating each filter with a binary mask modeled as an independent Bernoulli random variable, which can be continuously parameterized by the probability. Next, inspired by the recent work \cite{zhou2021effective}, we globally control the network size during the whole training process by controlling the sum of the Bernoulli distribution parameters. Thus, we can formulate the sparse network training problem into a constrained minimization problem on both the weights and structure parameters (i.e., the probability). The main novelty and contribution of this paper lies in our efficient training method called \textit{completely sparse neural network training} for solving the minimization problem. Specifically, to fully exploit the sparse structure, we separate training iteration into two parts, i.e., weight update and structure parameter update. For weight update, the conventional backward propagation is used to calculate the gradient, which can be sparsified completely because the gradients of the filters with zero valued masks are also zero. For structure parameter update, we develop a new \textbf{v}ariance \textbf{r}educed  \textbf{p}olicy \textbf{g}radient 
\textbf{e}stimator (VR-PGE). Unlike the conventional chain rule based gradient estimators (e.g., straight through\cite{bengio2013estimating}), VR-PGE estimates the gradient via two forward propagations, which is completely sparse because of the sparse subnetwork. Finally, extensive empirical results demonstrate that our method can significantly accelerate the training process of neural networks.  
 
 The main contributions of this paper can be summarized as follows:
 \begin{itemize}
     \item We develop an efficient sparse neural network training algorithm with the following three appealing features:
     \begin{itemize}
         \item In our algorithm, the computation in both forward and backward propagations is completely sparse, i.e., they do not need to go through any pruned channels, making the computational complexity significantly lower than that in standard training.
         \item During the whole training procedure, our algorithm works on small sub-networks with the {\it target sparsity} instead of follows a dense-to-sparse scheme.
         \item Our algorithm can be implemented easily on widely-used platforms, e.g., Pytorch and Tensorflow, to achieve practical speedup.
     \end{itemize}  
     \item We develop a variance reduced policy gradient estimator VR-PGE specifically for sparse neural network training, and prove that its variance is bounded.
     \item Experimental results demonstrate that our methods can achieve significant speed-up in training sparse neural networks. This implies that our method can enable us to explore larger-sized neural networks in the future.
 \end{itemize}

  \begin{remark}\label{remark-speedup}
 We call a sparse training algorithm  \textbf{completely sparse} if both its forward and backward propagation do not need to go through any pruned channels. For such algorithms, the computational cost in forward and backward propagation cost can be roughly reduced to $\rho^2 *100\%$, with $\rho$ being the ratio of remaining unpruned channels.
  \end{remark}

\section{Related Work}\label{sec:related-work}
In this section, we briefly review the studies on neural network pruning, which refers to the algorithms that prune DNNs after fully trained, and the recent works on sparse neural network training.  
\subsection{Neural Network Pruning}
Network Pruning \cite{han2015learning} is a promising technique for reducing the model size and inference time of DNNs. The key idea of existing methods \cite{han2015learning, guo2016dynamic,zeng2018mlprune,li2016pruning,luo2017thinet,he2017channel,zhu2017prune, wang2019structured, ye2020good,renda2020comparing, kang2020operation} is to develop effective criteria (e.g, weight magnitude) to identify and remove the massive unimportant weights contained in networks after training. To achieve practical speedup on general devices, some of them prune  networks in a structured manner, i.e., remove the weights in a certain group (e.g., filter) together, while others prune the weights individually. It has been reported in the literature \cite{guo2016dynamic,liu2018rethinking,zeng2018mlprune,li2016pruning} that they can improve inference efficiency and reduce memory usage of DNNs by orders of magnitudes with minor loss in accuracy, which enables the deployment of DNNs on low-power devices.

We notice that although some pruning methods  can be easily extended to train sparse networks,  they cannot accelerate or could even slow down the training process. One reason is they are developed in the scenario that a fully trained dense network is given, and cannot work well  on the  models learned in the early stage of training.  Another reason is  after each pruning iteration, one has to fine tune or even retrain the network for lots of epoch to  compensate the caused accuracy degradation.  
\subsection{Sparse Neural Network Training}
The research on sparse neural network training has emerged in the recent years. Different from the pruning methods, they can find sparse networks without pre-training a dense one.  Existing works can be divided into four categories based on their granularity in pruning and whether the network structures are explicitly parameterized.  To the best of our knowledge, no significant training speedups achieved in practice are reported in the literature.  Table \ref{table:summary-saprse-training} summarizes some representative works.  

\begin{table}[htb!]
\caption{Some representative works in sparse neural network training.}
\begin{center}
{\footnotesize
		\begin{tabular}{c|cc}
			\hline\hline
				granularity&non-parametric & parametric\\ \hline\hline
					\makecell*[c]{weight-level} & \makecell*[c]{\cite{menickrigging,frankle2019stabilizing,zhu2017prune, liu2015sparse,kusupati2020soft,mocanu2018scalable,wang2020picking,mostafa2019parameter,dettmers2019sparse}}& \makecell*[c]{\cite{xiao2019autoprune, srinivas2017training,louizos2017learning,zhou2021effective,kusupati2020soft}}\\\hline
					\makecell*[c]{channel-level}&\makecell*[c]{ \cite{wen2016learning,he2018soft}}& \makecell*[c]{\cite{lemaire2019structured,liu2017learning,yuan2020growing,louizos2017learning,kang2020operation}}\\ \hline\hline
			\end{tabular}
			}
\end{center}
	\label{table:summary-saprse-training}
\end{table}	
Weight-level non-parametric methods, e.g., \cite{frankle2019stabilizing,han2015learning,zhu2017prune, mocanu2018scalable, mostafa2019parameter}, always adopt a two-stage training procedure that alternates between weight optimization and pruning. They differ in the schedules of tuning the prune ratio over training and layers.  \cite{han2015learning} prunes the weights with the magnitude below a certain threshold and  \cite{zhu2017prune, frankle2019stabilizing} gradually increase the pruning rate during training.  \cite{mostafa2019parameter,menickrigging} automatically reallocate parameters across layers during training via controlling the global sparsity. 

Channel-level non-parametric methods \cite{he2018soft, wen2016learning} are proposed to achieve a practical acceleration in inference. \cite{ wen2016learning} is  a structured sparse learning method, which adds a group Lasso regularization into the objective function of DNNs with each group comprised of the weights in a filter.    \cite{he2018soft} proposes a soft filter pruning method. It zeroizes instead of hard pruning the filters with small $\ell_2$ norm, after which these filters are treated the same with other filters in training. It is obvious that these methods cannot achieve significant speedup in training since they need to calculate the full gradient in backward propagation although the forward propagation could be sparsified if implemented carefully. 

Parametric methods multiply each weight/channel with a binary \cite{zhou2021effective,yuan2020growing,srinivas2017training,xiao2019autoprune} or continuous \cite{liu2017learning,louizos2017learning, lemaire2019structured,kusupati2020soft} mask, which  can be either deterministic \cite{liu2017learning,xiao2019autoprune} or stochastic \cite{zhou2021effective,yuan2020growing,louizos2017learning,srinivas2017training, lemaire2019structured,kusupati2020soft}. The mask is always parameterized via a continuous trainable variable, i.e., structure parameter. The sparsity is achieved by adding sparsity-inducing regularizers on the masks. The novelties of these methods lie in estimating the gradients w.r.t structure parameters in training. To be precise,  
\begin{itemize}
    \item {\it Deterministic Binary Mask.} \cite{xiao2019autoprune} parameterizes its deterministic binary mask as a simple step function and estimates the gradients via sigmoid straight through estimator (STE) \cite{bengio2013estimating}.
    \item {\it Deterministic Continuous Mask.} \cite{liu2017learning} uses the linear coefficients of batch normalization (BN) as a continuous mask and enforces most of them to 0 by penalizing the objective with $\ell_1$ norm of the coefficients. \cite{kusupati2020soft} defines the mask as a  soft threshold function with learnable threshold. These methods can estimate the gradients via standard backward propagation. 
    \item {\it Stochastic Binary Mask.} \cite{yuan2020growing,srinivas2017training} model the mask as a bernoulli random variable and the gradients w.r.t. the parameters of bernoulli distributions are estimated via STE. \cite{zhou2021effective} estimates the gradients via Gumbel-Softmax trick \cite{jang2016categorical}, which is more accurate than STE. 
\item {\it Stochastic Continuous Mask.} \cite{louizos2017learning, lemaire2019structured} parameterize the mask as a continuous function $g(c, \epsilon)$, which is differentiable w.r.t. $c$, and $\epsilon$ is a parameter free noise, e.g., Gaussian noise $\mathcal{N}(0,1)$. In this way, the gradients can be calcuated via conventional backward propagation. 
\end{itemize}      Therefore, we can see that all of these parametric methods estimate the gradients of the structure parameters based on the chain rule in backward propagation. This makes the training iteration cannot be sparsified by exploiting the sparse network structure. For the details, please refer to Section \ref{sec:discussion}.

\begin{figure}[t!]
\centering  
\begin{tabular}{c}
\includegraphics[width=0.65\linewidth]{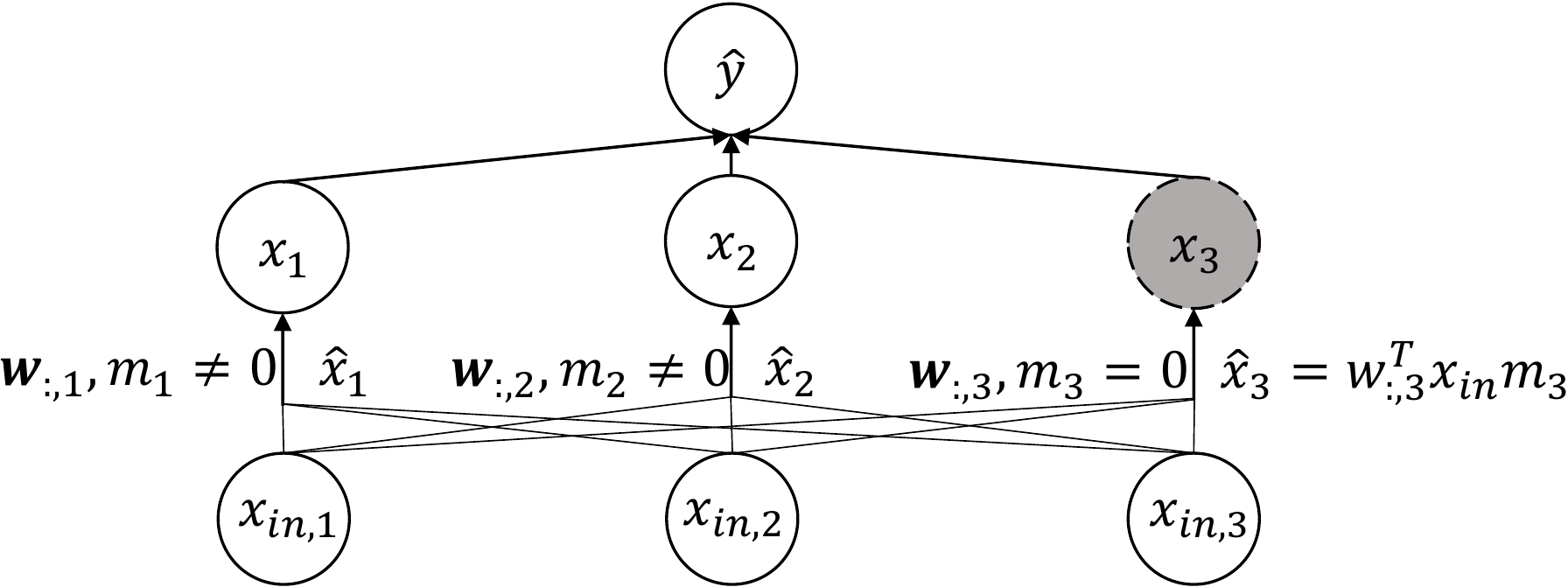}
\end{tabular}
\caption{A fully connected network. $\boldsymbol{w}$ is the weight matrix of $1$st layer, $m_i$ is the mask of $i$-th neuron; $\hat{y}$, $\hat{\x}_{in}$ and $\hat{x}_i$  are the output, input and preactivation. The $3$rd neuron (in grey) is pruned. }\label{fig:toy-network}
\end{figure}

\section{Why Existing Parameteric Methods Cannot Achieve Practical Speedup?}\label{sec:discussion}
In  this section, we reformulate existing  parametric channel-level methods into a unified framework to explain why they cannot accelerate the training process in practice. 

Notice that convolutional layer can be viewed as a generalized fully connected layer, i.e., viewing the channels as  neurons and convolution of two matrices as a generalized multiplication (see \cite{gu2020characterize}). Hence, for simplicity,  we consider the fully connected network in Figure \ref{fig:toy-network}. Moreover, since the channels in CNNs are corresponding to the neurons in fully connected networks, we consider {\it neuron-level instead of weight-level} sparse training in our example.  

As discussed in Section \ref{sec:related-work}, existing methods parameterize the 4 kinds of mask in the following ways:
\begin{align}
    \textup{(i): } m_i = \phi(s_i); \quad & \textup{(ii): } m_i = \psi(s_i); \quad  \textup{(iii): }m_i =g(s_i,\epsilon), \epsilon \sim \mathcal{N}(0,1); \quad  \textup{(iv): } m \sim \textup{Bern}(p_i(s)), \nonumber
\end{align}
where the function $\phi(s_i)$ is binary, e.g., step function; $\psi(s_i)$ is a  continuous function; $g(s_i,\epsilon)$ is differentiable w.r.t. $s_i$. All  existing methods estimate the gradient of the loss $\ell(\hat{y}, y)$ w.r.t. $s_i$ based on chain rule, which can be formulated into a unified form below. 

Specifically, we take the pruned neuron $x_3$ in Figure \ref{fig:toy-network} as an example, the gradient is calculated as   
\begin{align}
    \nabla_{s_3} \ell(\hat{y},y) =  \underbrace{\frac{\partial{\ell(\hat{y},y)}}{\partial \hat{x}_3 }}_{a} \underbrace{\left(\boldsymbol{w}^{\top}_{:,3} \x_{in}\right)}_{forward} \frac{\partial m_3}{\partial s_3}. \label{eqn:backward}
\end{align}
Existing parametric methods developed different ways to estimate $\frac{\partial m_3}{\partial s_3}$. Actually, for cases (ii) and (iii), the gradients are well-defined and thus can be calculated directly. STE is used to estimate the gradient in case (i)  \cite{xiao2019autoprune}. For cases (iv), \cite{yuan2020growing,srinivas2017training, zhou2021effective} adopt STE and Gumbel-Softmax.

In Eqn.(\ref{eqn:backward}), the term (a) is always nonzero especially when $\hat{x}_3$ is followed by BN. Hence, we can see that even for the pruned neuron $x_3$, the gradient  $\frac{\partial m_3}{\partial s_3}$ can be nonzero in all four cases. This means the backward propagation has to go though all the neurons/channels,  leading to dense computation. 

At last, we can know from Eqn.(\ref{eqn:backward}) that forward propagation in existing methods cannot be completely sparse. Although $\boldsymbol{w}^{\top}_{:,3} \x_{in}$ can  be computed  sparsely as in general models $\x_{in}$ could be a sparse tensor of a layer with some channels being pruned, we need to calculate it for {\it each} neuron via forward propagation to calculate RHS of Eqn.(\ref{eqn:backward}).  Thus, even if carefully implemented, the computational cost of forward propagation  can only be reduced to $\rho*100\%$ instead of $\rho^2*100\%$ as in inference. 

That's why we argue that existing methods need dense computation at least in backward propagation. So they cannot speed up the training process effectively in practice.

\begin{remark}
The authors of GrowEfficient \cite{yuan2020growing} confirmed that actually they also calculated the gradient of $q_c$ w.r.t, $\boldsymbol{s}_c$ in their Eqn.(6) via STE even if  $q_c=0$. Thus need dense backward propagation.
\end{remark}

\section{Channel-level Completely Sparse Neural Network Training}
Below, we present our sparse neural network training framework and the efficient training algorithm.
\subsection{Framework of Channel-level Sparse Training}
 Given a convolutional network $f(x;\boldsymbol{w})$, let $\{\calF_c : c \in \calC \}$ be the set of  filters with $\calC$ being the set of indices of all the channels. To parameterize the network structure,  we associate each $\calF_c$ with a binary mask $m_c$, which is an independent Bernoulli random variable. Thus, each channel is computed as 
$$
{\small
\boldsymbol{x}_{\text {out, c}}=\boldsymbol{x}_{i n}*\left(\mathcal{F}_cm_c\right),
}
$$
with $*$ being the convolution operation. Inspired by \cite{zhou2021effective}, to avoid the problems, e.g., gradient vanishing, we parameterize $m_c$ directly on the probability $s_c$, i.e., $m_c$ equals to 1 and 0 with the probabilities $s_c$ and $1-s_c$, respectively. Thus, we can control the channel size by the sum of $s_c$. Following \cite{zhou2021effective}, we can formulate  channel-level sparse network training into the following framework:
\begin{gather}
\min_{\boldsymbol{w}, \boldsymbol{s}} ~\mathbb{E}_{ p(\boldsymbol{m}|\boldsymbol{s})} ~\mathcal{L}(\boldsymbol{w}, \boldsymbol{m}):=\frac{1}{N}\sum_{i=1}^{N} \ell\left(f\left(\mathbf{x}_{i} ;\boldsymbol{w}, \boldsymbol{m} \right), \mathbf{y}_{i}\right) \label{eqn:continuous}\\
s.t.~ \boldsymbol{w}\in \mathbb{R}^n,  \boldsymbol{s}\in \mathcal{S} :=\{ \boldsymbol{s}\in [0,1]^{|\mathcal{C}|}: \boldsymbol{1}^\top \boldsymbol{s}  \leq K\}, \nonumber 
\end{gather}
where  $\left\{\left(\mathbf{x}_{i}, \mathbf{y}_{i}\right)\right\}_{i=1}^N$ is the training dataset, $\boldsymbol{w}$ is the weights of the original network,
$f\left(\cdot; \cdot, \cdot\right)$ is the pruned network, and $\ell(\cdot, \cdot)$ is the loss function, e.g, cross entropy loss.  $K = \rho|\mathcal{C}|$ controls the remaining channel size with $\rho$ being the remaining ratio of the channels. 

\textbf{Discussion.} We'd like to point out that although our  framework is inspired by  \cite{zhou2021effective}, our main contribution is the  efficient solver comprised of completely sparse forward/backward propagation for Problem (\ref{eqn:continuous}).  Moreover, our framework can  prune the weights in fully connected layers together, since we can associate each weight with an independent mask.

\subsection{Completely Sparse Training with Variance Reduced Policy Gradient}
Now we present our completely sparse training method, which can solve Problem (\ref{eqn:continuous}) via {\it completely} sparse forward and backward propagation. The key idea is to separate the training iteration into filter update and structure parameter update so that the sparsity can be fully exploited. 

\subsubsection{Filter Update via Completely Sparse Computation}
It is easy to see  that the computation of the gradient w.r.t. the filters can be sparsified completely. To prove this point, we just need to  clarify the following two things:
\begin{itemize}
    \item {\it We do not need to update the filters corresponding to the pruned channels.} Consider a pruned channel $c$, i.e., $m_c=0$, then due to the chain rule,  we can  have
\begin{align*}
    \frac{\partial \ell\left(f\left(\mathbf{x}_i;\boldsymbol{w}, \boldsymbol{m}\right)\right)}{\partial \mathcal{F}_c}= \frac{\partial \ell\left(f\left(\mathbf{x}_i;\boldsymbol{w}, \boldsymbol{m}\right)\right)}{\partial \boldsymbol{x}_{out,c}} \frac{\partial \boldsymbol{x}_{out,c}}{\partial \mathcal{F}_c}\equiv 0,
\end{align*}
the last equation holds since $\boldsymbol{x}_{out,c} \equiv 0$. This indicates that the gradient w.r.t the pruned filter $\mathcal{F}_c$ is always 0, and thus $\mathcal{F}_c$ does not need to be updated.  
\item {\it The error cannot pass the pruned channels via backward propagation.}  Consider a pruned channel $c$, we denote its output before masking as $\hat{\boldsymbol{x}}_{out, c} = \boldsymbol{x}_{i n}*\mathcal{F}_c$, then the error propagating through this channel can be computed as   
\begin{align*}
    \frac{\partial \ell\left(f\left(\mathbf{x}_i;\boldsymbol{w}, \boldsymbol{m}\right)\right)}{\partial \hat{\boldsymbol{x}}_{out, c}}= \frac{\partial \ell\left(f\left(\mathbf{x}_i;\boldsymbol{w}, \boldsymbol{m}\right)\right)}{\partial \boldsymbol{x}_{out,c}} \frac{\partial \boldsymbol{x}_{out,c}}{\hat{\boldsymbol{x}}_{out, c}}\equiv 0.
\end{align*}
This demonstrates that to calculate the gradient w.r.t. the  unpruned filters, the backward propagation does not need to go through  any pruned channels. 
\end{itemize} 
Therefore, the filters can be updated via  completely sparse  backward propagation. 

\subsubsection{Structure Parameter Update via Variance Reduced Policy Gradient}\label{sec:VR-PGE}

We notice that \textbf{p}olicy \textbf{g}radient \textbf{e}stimator (PGE) can estimate the gradient via forward propagation, avoiding the pathology of chain rule based estimators as dicussed in Section \ref{sec:discussion}. For abbreviation, we denote $\mathcal{L}(\boldsymbol{w}, \boldsymbol{m})$ as $\mathcal{L}(\boldsymbol{m})$ since $\boldsymbol{w}$ can be viewed as a constant here. The objective can be written as   
\begin{align*}
    \Phi(\boldsymbol{s}) = \mathbb{E}_{ p(\boldsymbol{m}|\boldsymbol{s})} ~\mathcal{L}(\boldsymbol{m}),
\end{align*}
which can be optimized using  gradient descent:
\[
\boldsymbol{s} \gets \boldsymbol{s} - \eta \nabla \Phi(\boldsymbol{s}).
\]
with   learning rate $\eta$. One can obtain a stochastic unbiased estimate of the gradient $\nabla \Phi(\boldsymbol{s})$ using  PGE:   
\begin{align}
   \nabla \Phi(\boldsymbol{s}) =  \mathbb{E}_{ p(\boldsymbol{m}|\boldsymbol{s})} ~\mathcal{L}(\boldsymbol{m}) \nabla_{\boldsymbol{s}} \ln{p(\boldsymbol{m}|\boldsymbol{s})} \tag{PGE} , \label{eqn:PGE}
\end{align}
leading to Policy Gradient method, which may be regarded as a stochastic gradient descent algorithm:
\begin{align}
\boldsymbol{s} \gets \boldsymbol{s} - \eta \mathcal{L}(\boldsymbol{m}) \nabla_{\boldsymbol{s}} \ln{p(\boldsymbol{m}|\boldsymbol{s})}.\label{eqn:update-PGE}
\end{align}
In Eqn.(\ref{eqn:update-PGE}), $\mathcal{L}(\boldsymbol{m})$ can be computed via completely sparse forward propagation and the computational cost of $\nabla_{\boldsymbol{s}} \ln{p(\boldsymbol{m}|\boldsymbol{s})} = \frac{\boldsymbol{m}- \boldsymbol{s}}{\boldsymbol{s}(1-\boldsymbol{s})}$ is negligible, therefore PGE is computationally efficient. 

However, in accordance with the empirical results reported in \cite{rezende2014stochastic,jang2016categorical}, we found that standard PGE suffers from high variance and does not work in practice. Below we will develop a \textbf{V}ariance \textbf{R}educed \textbf{P}olicy \textbf{G}radient \textbf{E}stimator (VR-PGE) starting from theoretically analyzing the variance of PGE.

Firstly, we know that this variance of PGE is
\begin{align}
     \mathbb{E}_{ p(\boldsymbol{m}|\boldsymbol{s})} ~\mathcal{L}^2(\boldsymbol{m}) \|\nabla_{\boldsymbol{s}} \ln{p(\boldsymbol{m}|\boldsymbol{s})}\|_2^2 - \|\nabla \Phi(\boldsymbol{s})\|_2^2, \nonumber
\end{align}
which can be large because $\mathcal{L}(\boldsymbol{m})$ is large. 

Mean Field theory \cite{song2018mean} indicates that, while  $\mathcal{L}(\boldsymbol{m})$ can be large, the term $\mathcal{L}(\boldsymbol{m}) - \mathcal{L}(\boldsymbol{m}')$ is small when $\boldsymbol{m}$ and $\boldsymbol{m'}$  are two independent masks sampled from a same distribution$ p(\boldsymbol{m}|\boldsymbol{s})$ (see the appendix for the details). This means that we may consider the following variance reduced preconditioned policy gradient estimator:
\begin{align}
 \mathbb{E}_{\boldsymbol{m}' \sim  p(\boldsymbol{m}'|\boldsymbol{s})} \mathbb{E}_{\boldsymbol{m}\sim  p(\boldsymbol{m}|\boldsymbol{s})} ~\left(\mathcal{L}(\boldsymbol{m})- \mathcal{L}(\boldsymbol{m}')\right)H^{\alpha}(\boldsymbol{s}) \nabla_{\boldsymbol{s}} \ln{p(\boldsymbol{m}|\boldsymbol{s})} \tag{VR-PGE}, \label{eqn:VR-PGE}
\end{align}
where $H^{\alpha}(\boldsymbol{s})$ is  a specific diagonal preconditioning matrix 
\begin{equation}
H^{\alpha}(\boldsymbol{s}) = \textup{diag}\left(\boldsymbol{s}\circ(1- \boldsymbol{s})\right)^{\alpha},
\label{eq:H}
\end{equation}
with $\alpha \in (0,1)$ and $\circ$ being the element-wise product. It plays a role as adaptive step size and it is shown that this term can reduce the variance of the stochastic PGE term $\nabla_{\boldsymbol{s}}\ln{p(\boldsymbol{m}|\boldsymbol{s})}$. The details can be found in the appendix. Thus $\Phi(\s)$ can be optimized via:
\begin{align}
\boldsymbol{s} \gets \boldsymbol{s} - \eta \left(\mathcal{L}\left(\boldsymbol{m}\right)- \mathcal{L}\left(\boldsymbol{m}'\right)\right)H^{\alpha}(\boldsymbol{s}) \nabla_{\boldsymbol{s}} \ln{p(\boldsymbol{m}|\boldsymbol{s})}.\label{eqn:update-VR-PGE}
\end{align}
In our experiments, we set $\alpha$ to be $\frac{1}{2}$ for our estimator \ref{eqn:VR-PGE}. The theorem below demonstrates that  \ref{eqn:VR-PGE} can have bounded variance. 
\begin{theorem}\label{thm:variance-bound}
Suppose $\boldsymbol{m}$ and $\boldsymbol{m'}$  are two independent masks sampled from the Bernoulli distribution $ p(\boldsymbol{m}|\boldsymbol{s})$, then for any $\alpha\in [\frac{1}{2}, 1)$ and $\bs{s}\in (0,1)^{|\mathcal{C}|}$, the variance is bounded for
\begin{align*}
\left(\mathcal{L}(\boldsymbol{m})- \mathcal{L}(\boldsymbol{m}')\right)H^{\alpha}(\boldsymbol{s}) \nabla_{\boldsymbol{s}} \ln{p(\boldsymbol{m}|\boldsymbol{s})}
\end{align*}
\end{theorem}

Finally, we provide a complete view of our  sparse training algorithm in Algorithm \ref{alg:FBPS}, which is essentially a projected stochastic gradient descent equipped with our efficient gradient estimators above. The projection operator in Algorithm \ref{alg:FBPS} can be computed efficiently using Theorem 1 of \cite{zhou2021effective}.  
\begin{tiny}
\begin{algorithm*}[t!]
\caption{Completely Sparse Neural Network Training}
\label{alg:FBPS}
\begin{algorithmic}[1]
\REQUIRE target remaining ratio $\rho$, a dense network $\bs{w}$, the step size $\eta$, and parameter $\alpha$ in \eqref{eq:H} .
\STATE Initialize $\bs{w}$, let $\bs{s}=\rho \mathbf{1}$.
\FOR {training epoch $t = 1, 2 \ldots T$}
\FOR {each training iteration}
\STATE Sample mini batch of data $\mathcal{B} = \left\{\left(\mathbf{x}_{1}, \mathbf{y}_{1}\right), \ldots,\left(\mathbf{x}_{B}, \mathbf{y}_{B}\right)\right\}$.
\STATE Sample $\boldsymbol{m}^{(i)}$ from $p(\boldsymbol{m}|\boldsymbol{s})$, $ i=1, 2$.
\STATE Update $\bs{s}$ and $\bs{w}$ \\
$\bs{s} \leftarrow \operatorname{proj}_{\mathcal{S}}(\bs{z}) \mbox{ with }\bs{z} = \bs{s}-\eta \left(\mathcal{L}_{\mathcal{B}}(\boldsymbol{w}, \bs{m}^{(1)})- \mathcal{L}_{\mathcal{B}}(\boldsymbol{w}, \bs{m}^{(2)})\right) H^{\alpha}(\bs{s})\frac{\bs{m}^{(1)}-\bs{s}}{\bs{s}(1-\bs{s})}, \label{eqn:proj}$\\
$\bs{w} \leftarrow \bs{w} -   \eta \nabla_{\boldsymbol{w}} \mathcal{L}_{\mathcal{B}}\left(\boldsymbol{w}, \bs{m}^{(1)}\right)\nonumber$
\ENDFOR
\ENDFOR

\RETURN A pruned network $\bs{w}\circ \boldsymbol{m}$ by sampling a mask $\bs{m}$ from the distribution $p(\boldsymbol{m}|\boldsymbol{s})$.
\end{algorithmic}
\end{algorithm*}
\end{tiny}

\textbf{Discussion.} In our algorithm, benefited from our constraint on $\bs{s}$, the channel size of the neural network during training can be strictly controlled. This is in contrast with GrowEfficient \cite{yuan2020growing}, which ultilizes regularizer term to control the model size and has situations where model size  largely drift away from desired. This will have larger demand for the GPU memory storage and have more risk that memory usage may explode, especially when we utilize sparse learning to explore larger models. Moreover, our forward and backward propagations are completely sparse, i.e., they do not need to go through any pruned channels. Therefore, the computational cost of each training iteration can be roughly reduced to $\rho^2*100\%$ of the dense network.

\section{Experiments}\label{sec:exp}
In this section, we conduct a series of experiments to demonstrate the outstanding performance of our method. We divide the experiments into five parts. In part one, we compare our method with several state-of-the-art methods on CIFAR-10 \cite{krizhevsky2009learning} using VGG-16 \cite{simonyan2014very}, ResNet-20 \cite{he2016deep} and WideResNet-28-10 \cite{wideresnet} to directly showcase the superiority of our method. In part two, we directly compare with state-of-the-art method GrowEfficient \cite{yuan2020growing} especially on extremely sparse regions, and on two high capacity networks VGG-19 \cite{simonyan2014very} and ResNet-32 \cite{he2016deep} on CIFAR-10/100 \cite{krizhevsky2009learning}. In part three, we conduct experiments on a large-scale dataset ImageNet \cite{deng2009imagenet} 
with ResNet-50 \cite{he2016deep} and MobileNetV1 \cite{howard2017mobilenets} and compare with GrowEfficient \cite{yuan2020growing} across a wide sparsity region. In part four, we present the train-computational time as a supplementary to the conceptual train-cost savings to justify the applicability of sparse training method into practice. In part five, we present further analysis on epoch-wise train-cost dynamics and experimental justification of variance reduction of VR-PGE. Due to the space limitation, we postpone the experimental configurations, calculation schemes on train-cost savings and train-computational time and additional experiments into appendix. 

\subsection{VGG-16, ResNet-20 and WideResNet-28-10 on CIFAR-10}

\begin{table*}[t!]
\caption{Comparison with the channel pruning methods L1-Pruning \cite{li2016pruning}, SoftNet \cite{he2018soft}, ThiNet \cite{luo2017thinet}, Provable \cite{liebenwein2020provable} and one channel sparse training method GrowEfficient \cite{yuan2020growing} on CIFAR-10.}
\begin{center}
{\footnotesize
\begin{tabular}{p{1.65cm}<{\centering} p{1.65cm}<{\centering}p{1.65cm}<{\centering}p{1.65cm}<{\centering}p{1.65cm}<{\centering}p{3cm}<{\centering}}\toprule
Model& Method& Val Acc(\%) &Params(\%)& FLOPs(\%)& Train-Cost Savings($\times$)\\ \cmidrule(){1-6}
\multirow{7}{*}{VGG-16}& Original & 92.9 & 100 & 100 & 1$\times$ \\
      & L1-Pruning & 91.8 & 19.9 & 19.9 & - \\
      & SoftNet & 92.1 & 36.0 & 36.1 & - \\
      & ThiNet & 90.8 & 36.0 & 36.1 & - \\
      & Provable & 92.4 & 5.7 & 15.0 & - \\
      & GrowEfficient & 92.5 & 5.0 & 13.6 & 1.22$\times$ \\
      & Ours & \textbf{92.5} & \textbf{4.4} & \textbf{8.7} & \textbf{8.69}$\times$ \\ \cmidrule(){1-6}
\multirow{7}{*}{ResNet-20}& Original & 91.3 & 100 & 100 & 1$\times$ \\
         & L1-Pruning & 90.9 & 55.6 & 55.4 & - \\
         & SoftNet & 90.8 & 53.6 & 50.6 & - \\
         & ThiNet & 89.2 & 67.1 & 67.3 & - \\
         & Provable & 90.8 & 37.3 & 54.5 & - \\
         & GrowEfficient & 90.91 & 35.8 & 50.2 & 1.13$\times$ \\
         
         & Ours & \textbf{90.93} & \textbf{35.1} & \textbf{36.1} & \textbf{2.09}$\times$ \\
         \cmidrule(){1-6}

\multirow{4}{*}{WRN-28-10}& Original & 96.2 & 100 & 100 & 1$\times$ \\
         & L1-Pruning & 95.2 & 20.8 & 49.5 & - \\
         & GrowEfficient & 95.3 & 9.3 & 28.3 & 1.17$\times$ \\
         & Ours & \textbf{95.6} & \textbf{8.4} & \textbf{7.9} & \textbf{9.39}$\times$ \\
         \bottomrule
\end{tabular}
}
\end{center}
\label{tab:VGG19-CIFAR10-100}
\end{table*}

Table \ref{tab:VGG19-CIFAR10-100} presents Top-1 validation accuracy, parameters, FLOPs and train-cost savings comparisons with channel pruning methods L1-Pruning \cite{li2016pruning}, SoftNet \cite{he2018soft}, ThiNet \cite{luo2017thinet}, Provable \cite{liebenwein2020provable} and  sparse training method GrowEfficient \cite{yuan2020growing}. SoftNet can train from scratch but requires completely dense computation. Other pruning methods all require pretraining of dense model and multiple rounds of pruning and finetuning, which makes them slower than vanilla dense model training. Therefore the train-cost savings of these methods are below 1$\times$ and thus shown as ("-") in Table \ref{tab:VGG19-CIFAR10-100}.

GrowEfficient \cite{yuan2020growing} is a recently proposed state-of-the-art channel-level sparse training method showing train-cost savings compared with dense training. As described in Section \ref{sec:discussion}, GrowEfficient features completely dense backward and partially sparse forward pass, making its train-cost saving limited by $\frac{3}{2}$. By contrast, the train-cost savings of our method is not limited by any constraint. The details of how train-cost savings are computed can be found in appendix.

Table \ref{tab:VGG19-CIFAR10-100} shows that our method generally exhibits better performance in terms of validation accuracy, parameters and particularly FLOPs. In terms of train-cost savings, our method shows at least 1.85$\times$ speed-up against GrowEfficient \cite{yuan2020growing} and up to 9.39$\times$ speed-up against dense training.  

\subsection{Wider Range of Sparsity on CIFAR-10/100 on VGG-19 and ResNet-32}

\begin{figure}[t!]
\begin{center}
\centering  
\includegraphics[width=0.83\linewidth]{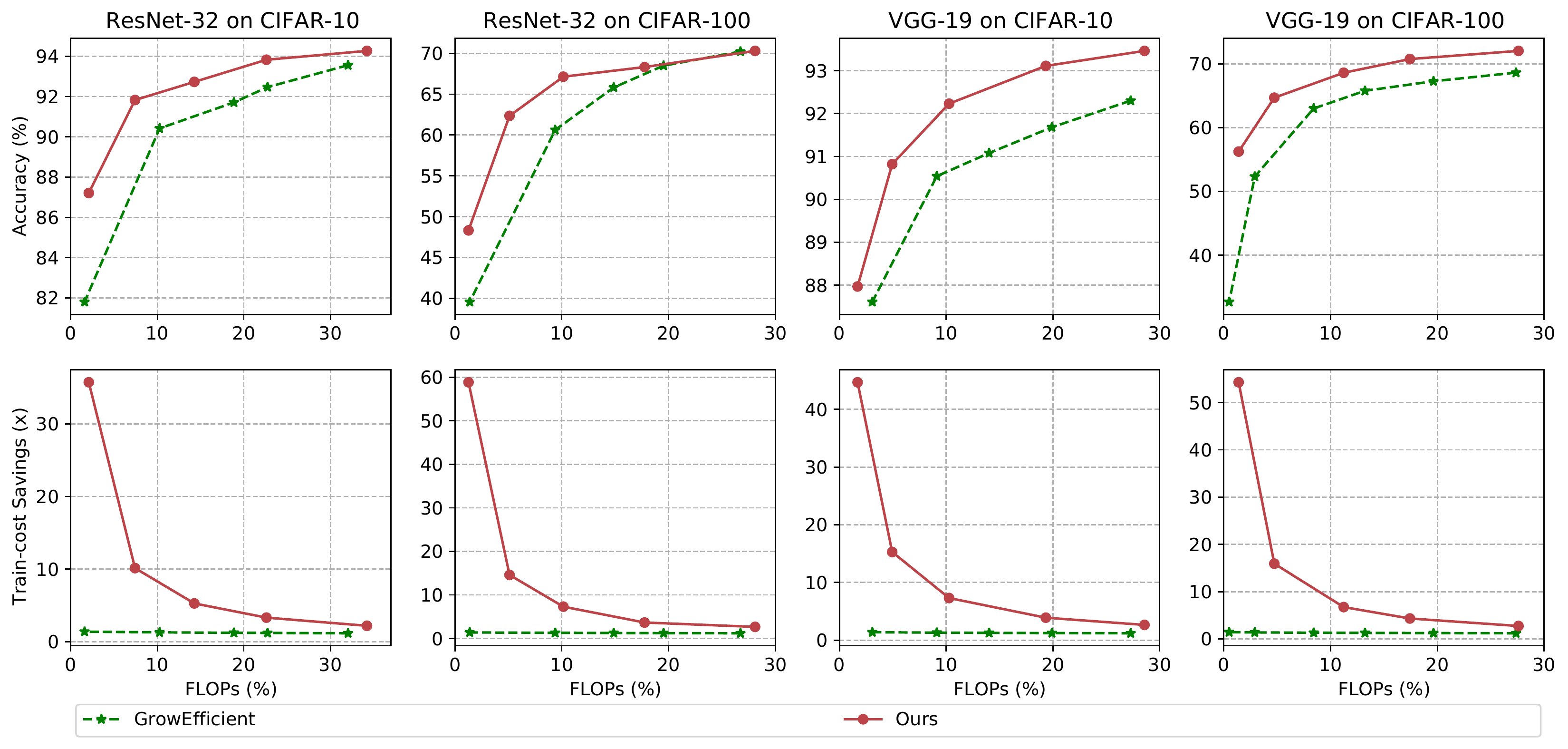}
\end{center}
\caption{Comparison of Top-1 Validation Accuracy and Train-cost Savings on CIFAR-10/100.}\label{fig:result-CIFAR-10-100}
\end{figure}

In this section, we explore sparser regions of training efficiency to present a broader comparision with state-of-the-art channel sparse training method GrowEfficient \cite{yuan2020growing}.

We plot eight figures demonstrating the relationships between the Top-1 validation accuracy, FLOPs and train-cost savings. We find that our method generally achieves higher accuracy under same FLOPs settings. To be noted, the train-cost savings of our method is drastically higher than GrowEfficient \cite{yuan2020growing}, reaching up to 58.8$\times$ when sparisty approches 1.56\% on ResNet-32 on CIFAR-100, while the speed-up of GrowEfficient is limited by $\frac{3}{2}$. 

\begin{table*}[t!]
\caption{Comparison with the channel pruning methods L1-Pruning \cite{li2016pruning}, SoftNet \cite{he2018soft}, Provable \cite{liebenwein2020provable} and one channel sparse training method GrowEfficient \cite{yuan2020growing} on ImageNet-1K.}
\begin{center}
{\footnotesize
\begin{tabular}{p{1.65cm}<{\centering} p{1.65cm}<{\centering}p{1.65cm}<{\centering}p{1.65cm}<{\centering}p{1.65cm}<{\centering}p{3cm}<{\centering}}\toprule
Model& Method& Val Acc(\%) &Params(\%)& FLOPs(\%)& Train-Cost Savings($\times$)\\ \cmidrule(){1-6}
\multirow{7}{*}{ResNet-50}& Original & 77.0 & 100 & 100 & 1$\times$ \\
      & L1-Pruning & 74.7 & 85.2 & 77.5 & - \\
      & SoftNet & 74.6 & - & 58.2 & - \\
      & Provable & 75.2 & 65.9 & 70.0 & - \\
      & GrowEfficient & 75.2 & 61.2 & 50.3 & 1.10$\times$ \\
      & Ours & \textbf{76.0} & 48.2 & 46.8 & 1.60$\times$ \\
      & Ours & 73.5 & 27.0 & 24.7 &  3.02$\times$ \\
      & Ours & 69.3 & \textbf{10.8} & \textbf{10.1} &  \textbf{7.36}$\times$\\
\bottomrule
\end{tabular}
}
\end{center}
\label{tab:ResNet50-ImageNet}
\end{table*}

\subsection{ResNet-50 and MobileNetV1 on ImageNet-1K}
\begin{figure}[t!]
\begin{center}
\centering  
\includegraphics[width=0.95\linewidth]{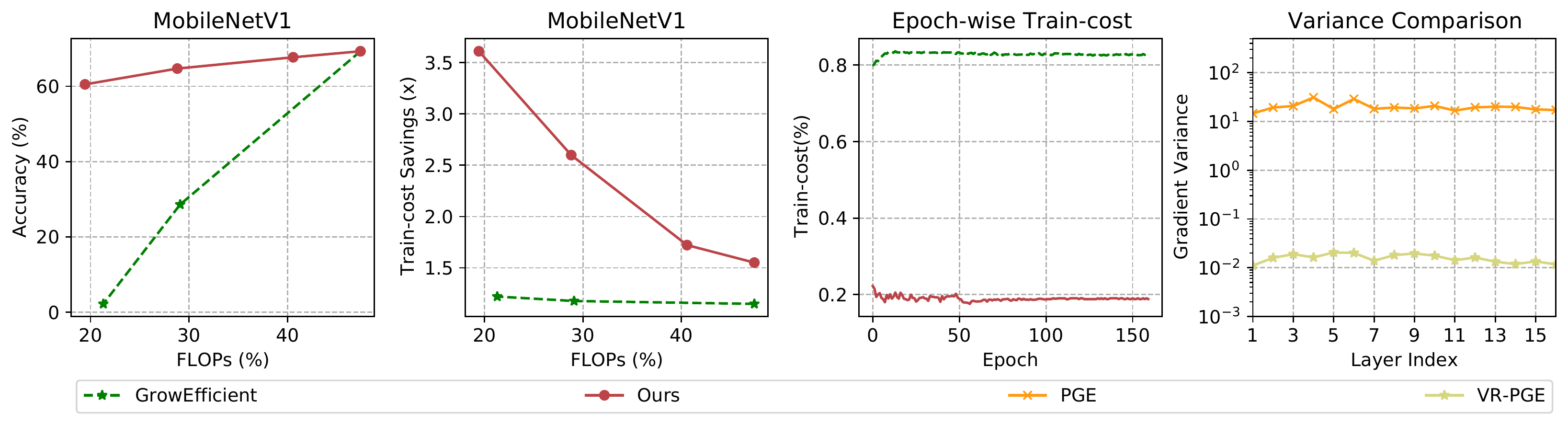}
\end{center}
\caption{Top-1 Validation Accuracy and Train-cost Savings on MobileNetV1 on ImageNet. Epoch-wise Train-cost and Variance Comparison on VGG-19 on CIFAR-10. }\label{fig:result-ImageNet}
\end{figure}

In this section, we present the performance boost obtained by our method on ResNet-50 and MobileNetV1 on  ImageNet-1K \cite{deng2009imagenet}. Our method searches a model with 76.0\% Top-1 accuracy, 48.2\% parameters and 46.8\% FLOPs beating all compared state-of-the-art methods. The train-cost saving comes up to 1.60$\times$ and is not prominent due to the accuracy constraint to match up with compared methods. Therefore we give a harder limit to the channel size and present sparser results on the same Table \ref{tab:ResNet50-ImageNet}, reaching up to 7.36$\times$ speed-up while still preserving 69.3\% Top-1 accuracy. For the already compact model MobileNetV1, we plot two figures in Figure \ref{fig:result-ImageNet} comparing with GrowEfficient \cite{yuan2020growing}. We find that our method is much stabler in sparse regions and obtains much higher train-cost savings. 




\subsection{Actual Training Computational Time Testing}
In this section, we provide actual training computational time on VGG-19 and CIFAR-10. The GPU in test is RTX 2080 Ti and the deep learning framework is Pytorch \cite{paszke2019pytorch}. The intent of this section is to justify the feasibility of our method in reducing actual computational time cost, rather than staying in conceptual training FLOPs reduction. The computational time cost is measured by wall clock time, focusing on forward and backward propagation. We present training computational time in Table \ref{tab:tct} with varying sparsity as in Figure \ref{fig:result-CIFAR-10-100}. It shows that the computational time savings increases steadily with the sparisty. We also notice the gap between the savings in FLOPS and computational time. The gap comes from the difference between FLOPs and actual forward/backward time. More specifically, forward/backward time is slowed down by data-loading processes and generally affected by hardware latency and throughput, network architecture, etc. At extremely sparse regions, the pure computational time of sparse networks only occupies little of the forward/backward time and the cost of data management and hardware latency dominates the wall-clock time. Despite this gap, it can be expected that our train-cost savings can be better translated into real speed-up in exploring large models where the pure computational time dominates the forward/backward time, which promises a bright future for making training infeasibly large models into practice.

\begin{table*}[htb!]
\caption{Train-computational Time on VGG-19 with CIFAR-10. The computational time saving is not as prominent as train-cost savings while still achieving nearly an order of reduction, preserving 87.97\% accuracy. }
\begin{center}
{\footnotesize
\begin{tabular}{p{1.65cm}<{\centering} p{1.65cm}<{\centering}p{1.65cm}<{\centering}p{1.65cm}<{\centering}p{1.65cm}<{\centering}p{3cm}<{\centering}}\toprule
Model& Val Acc(\%) &Params(\%)& FLOPs(\%)& Train-Cost Savings($\times$) & Train-Computational Time(min) \\ \cmidrule(){1-6}
\multirow{6}{*}{VGG-19} & 93.84 & 100.00 & 100.00 & 1.00$\times$ & 21.85 (1.00$\times$)\\
      & 93.46 & 23.71 & 28.57 & 2.64$\times$ & 14.04 (1.55$\times$)\\
      & 93.11 & 12.75 & 19.33 & 3.89$\times$ & 10.43 (2.09$\times$)\\
      & 92.23 & 6.69 & 10.27 & 7.30$\times$ & 6.83 (3.20$\times$)\\
      & 90.82 & 3.06 & 4.94 & 15.28$\times$ & 4.86 (4.50$\times$)\\
      & 87.97 & 0.80 & 1.70 & 44.68$\times$ & 2.95 (7.41$\times$)\\ 
      \bottomrule
\end{tabular}
}
\end{center}
\label{tab:tct}
\end{table*}




\subsection{Further Analysis}

\textbf{[Epoch-wise Train-cost Dynamics of Sparse Training Process]} We plot the train-cost dynamics in Figure \ref{fig:result-ImageNet}. The vertical label is the ratio of train-cost to dense training, the inverse of train-cost savings. This demonstrates huge difference between our method and GrowEfficient \cite{yuan2020growing}. The model searched by our method exhibits 92.73\% Top-1 accuracy, 16.68\% parameters, 14.28\% FLOPs with 5.28$\times$ train-cost savings, while the model searched by GrowEfficient exhibits 92.47\% Top-1 accuracy, 18.08\% parameters, 22.74\% FLOPs with 1.21$\times$ train-cost savings.

\textbf{[Experimental Verification of Variance Reduction of VR-PGE against PGE]} We plot  the mean of variance of gradients of channels from different layers. The model checkpoint and input data are selected randomly. The gradients are calculated in two approaches, VR-PGE and PGE. From the rightmost graph of Figure \ref{fig:result-ImageNet}, we find that the VR-PGE reduces variance significantly, up to 3 orders of magnitude.

\section{Conclusion}
This paper proposes an efficient sparse neural network training method with completely sparse forward and backward passes. A novel gradient estimator named VR-PGE is developed for updating structure parameters, which estimates the gradient via two sparse forward propagation.  We theoretically proved that VR-PGE has bounded variance. In this way, we can separate the weight and structure update in training and making the whole training process completely sparse.    Emprical results demonstrate that the proposed method can significantly accelerate the training process of DNNs in practice. This enables us to explore larger-sized neural networks in the future. 
\begin{ack}
    This work is supported by GRF 16201320.
\end{ack}

\clearpage
\newpage
\appendix
\newcommand{\mytoptitlebar}{
  \noindent\rule{\textwidth}{4pt}
  \vskip 0.16in
}
\newcommand{\mybottomtitlebar}{
  \noindent\rule{\textwidth}{1.pt}
}

\mytoptitlebar
\begin{center}
	{\Large\bf Supplemental Material: Efficient Neural Network Training via Forward and Backward Propagation Sparsification \par }
\end{center}
\mybottomtitlebar
\vskip 0.4in

This appendix can be divided into four parts. To be precise,
\begin{itemize}
    \item[1.] Section \ref{sec:proof-theorem1} gives the detailed proof of Theorem \ref{thm:variance-bound} and discuss the convergence of our method.
    \item[2.] Section \ref{sec:appendix-experiment} present experimental configurations of this work.
    \item[3.] Section \ref{sec:appendix-calculation} present calculation schemes on train-cost savings and train-computational time.
    \item[4.] Section \ref{sec:appendix-discuss} discusses the potentials and limitations of this work. 
\end{itemize}

\section{Proof of Theorem \ref{thm:variance-bound}}\label{sec:proof-theorem1}
\subsection{Properties of Overparameterized Deep Neural Networks}\label{sec:property}
Before giving the detailed proof, we would like to present the following two properties of overparameterized deep neural networks, which are implied by the latest studies based on the  mean field theory. We will empirically verify these properties in this section and  adopt them as assumptions in our  proof. 

\begin{property}\label{property:sampling}
Given the probability $\bs{s}$ and the weights $\bs{w}$ for an overparameterized deep neural network, then for two independent masks $\bs{m}$ and $\bs{m}'$ sampled from $p(\cdot| \bs{s})$, $\mathcal{L}(\bs{m})-\mathcal{L}(\bs{m}')$ is always small. That is  
\begin{align}
    V(\bs{s}):=\rE_{\bs{m}\sim p(\cdot|\bs{s})} \rE_{\bs{m}'\sim p(\cdot|\bs{s})} \left(\mathcal{L}(\boldsymbol{m})- \mathcal{L}(\boldsymbol{m}')\right)^2 \label{eqn:property-sampling}
\end{align}
is small. 
\end{property}
The mean field theory based studies \cite{song2018mean, fang2019convex} proved that discrete deep neural networks can be viewed as sampling neurons/channels from continuous networks according to certain distributions.  As the numbers of neurons/channels increase, the output of discrete networks would converge to that of the continuous networks (see Theorem 3 in \cite{song2018mean} and Theorem 1 in \cite{fang2019convex}). Although in standard neural networks we do not have the scaling operator as \cite{song2018mean,fang2019convex} for computing the expectation, due to the batch normalization layer, the affect caused by this difference can largely be eliminated. The subnetworks $\bs{m}$ and $\bs{m}'$ here can be roughly viewed as sampled from a common continuous network. Therefore, $\mathcal{L}(\boldsymbol{m})- \mathcal{L}(\boldsymbol{m}')$ would be always small. That's why Property \ref{property:sampling} holds. 

\begin{property}\label{property:flip}
Given the probability $\bs{s}$ and the weights $\bs{w}$ for an overparameterized deep neural network, consider a mask $\bs{m}$ sampled from $p(\cdot| \bs{s})$, if we flip one component of $\bs{m}$, then the network would not change too much. Combined with Property \ref{property:sampling}, this can be stated as:  for any $j\in \mathcal{C}$, we denote $\bs{m}_{-j}$ and $\bs{s}_{-j}$ to be all the components of $\bs{m}$ and $\bs{s}$ except the $j$-th component, and define
\begin{align*}
    V_{\max}(\bs{s}) := \max_{\bs{m}_j \in \{0,1\}, j\in \mathcal{C} }\rE_{\bs{m}_{-j}\sim p(\cdot|\bs{s}_{-j})} \rE_{\bs{m}'\sim p(\cdot|\bs{s})} \left(\mathcal{L}(\boldsymbol{m})- \mathcal{L}(\boldsymbol{m}')\right)^2,
\end{align*}
then 
\begin{align}
     V_{\max}(\bs{s}) \approx V(\bs{s}). \label{eqn:property-flip}
\end{align}
\end{property}
In the mean field based studies \cite{song2018mean,fang2019convex}, they model output of a neuron/channel as a expectation of weighted sum of the neurons/channels in the previous layer w.r.t. a certain distribution. Therefore, the affect of flipping one component of the mask on expectation is negligible.  Therefore Property \ref{property:flip} holds.    

\subsection{Detailed Proof}
\begin{proof}
In this proof, we denote 
\begin{align*}
\left(\mathcal{L}(\boldsymbol{m})- \mathcal{L}(\boldsymbol{m}')\right)H^{\alpha}(\boldsymbol{s}) \nabla_{\boldsymbol{s}} \ln{p(\boldsymbol{m}|\boldsymbol{s})}
\end{align*}
as $\mathcal{G}^{\alpha}(\bs{m}, \bs{m}'|\bs{s})$. Note that the total variance
\begin{align}
    &\mbox{Var} (\mathcal{G}^{\alpha}(\bs{m}, \bs{m}'|\bs{s}))\nonumber \\
    =& \rE_{\bs{m}\sim p(\cdot|\bs{s})} \rE_{\bs{m}'\sim p(\cdot|\bs{s})}\| \mathcal{G}^{\alpha}(\bs{m}, \bs{m}'|\bs{s})\|_2^2 -  \|\rE_{\bs{m}\sim p(\cdot|\bs{s})} \rE_{\bs{m}'\sim p(\cdot|\bs{s})} \mathcal{G}^{\alpha}(\bs{m}, \bs{m}'|\bs{s})\|_2^2,\nonumber
\end{align}
we only need to prove that the term $\rE_{\bs{m}\sim p(\cdot|\bs{s})} \rE_{\bs{m}'\sim p(\cdot|\bs{s})}\| \mathcal{G}^{\alpha}(\bs{m}, \bs{m}'|\bs{s})\|_2^2$ is bounded. 

We let $\bs{m}_{-j}$ and $\bs{s}_{-j}$ be all the components of $\bs{m}$ and $\bs{s}$ except the $j$-th component with $j\in \mathcal{C}$. We consider the $j$-th component of $\mathcal{G}^{\alpha}(\bs{m}, \bs{m}'|\bs{s})$, i.e., $\mathcal{G}_j^{\alpha}(\bs{m}, \bs{m}'|\bs{s})$, then  $\rE_{\bs{m}\sim p(\cdot|\bs{s})} \rE_{\bs{m}'\sim p(\cdot|\bs{s})}\| \mathcal{G}_j^{\alpha}(\bs{m}, \bs{m}'|\bs{s})\|_2^2$ can be estimated as
\begin{align}
    &\rE_{\bs{m}\sim p(\cdot|\bs{s})} \rE_{\bs{m}'\sim p(\cdot|\bs{s})} \left(\mathcal{G}_j^{\alpha}(\bs{m}, \bs{m}'|\bs{s})\right)^2\nonumber \\
    =& \rE_{\bs{m}\sim p(\cdot|\bs{s})} \rE_{\bs{m}'\sim p(\cdot|\bs{s})} \left(\mathcal{L}(\boldsymbol{m})- \mathcal{L}(\boldsymbol{m}')\right)^2 [H^{\alpha}(\boldsymbol{s}) \nabla_{\boldsymbol{s}} \ln{p(\boldsymbol{m}|\boldsymbol{s})}]_j^2\nonumber\\
    =& \rE_{\bs{m}\sim p(\cdot|\bs{s})} \rE_{\bs{m}'\sim p(\cdot|\bs{s})} \left(\mathcal{L}(\boldsymbol{m})- \mathcal{L}(\boldsymbol{m}')\right)^2 \left( \bs{s}_{j}^{2\alpha}(1-\bs{s}_j)^{2\alpha} \frac{(\bs{m}_j-\bs{s}_j)^2}{\bs{s}^2_j(1-\bs{s}_j)^2}\right)\label{eqn:proof-2} \\ 
    =&\rE_{\bs{m}\sim p(\cdot|\bs{s})} \rE_{\bs{m}'\sim p(\cdot|\bs{s})} \left(\mathcal{L}(\boldsymbol{m})- \mathcal{L}(\boldsymbol{m}')\right)^2 \left( \bs{s}_{j}^{2(\alpha-1)}(1-\bs{s}_j)^{2(\alpha-1)} (\bs{m}_j-\bs{s}_j)^2\right)\nonumber \\
    =&\rE_{\bs{m}_{j}\sim p(\cdot|\bs{s}_{j})}\left(\rE_{\bs{m}_{-j}\sim p(\cdot|\bs{s}_{-j})} \rE_{\bs{m}'\sim p(\cdot|\bs{s})} \left(\mathcal{L}(\boldsymbol{m})- \mathcal{L}(\boldsymbol{m}')\right)^2 \right) \left( \bs{s}_{j}^{2(\alpha-1)}(1-\bs{s}_j)^{2(\alpha-1)} (\bs{m}_j-\bs{s}_j)^2\right)\nonumber\\
    \mathop{\leq}_{\eqref{eqn:property-flip}}& V_{\max}(\bs{s}) \rE_{\bs{m}_{j}\sim p(\cdot|\bs{s}_{j})}\left( \bs{s}_{j}^{2(\alpha-1)}(1-\bs{s}_j)^{2(\alpha-1)} (\bs{m}_j-\bs{s}_j)^2\right)\label{eqn:proof-3} \\
    =& \left( \bs{s}_j^{2\alpha}(1-\bs{s}_j)^{(2\alpha-1)}+ \bs{s}_j^{2\alpha-1}(1-\bs{s}_j)^{2\alpha}\right)V_{\max}(\bs{s}) \label{eqn:proof-1}.
\end{align}
Thus $\rE_{\bs{m}\sim p(\cdot|\bs{s})} \rE_{\bs{m}'\sim p(\cdot|\bs{s})}\| \mathcal{G}^{\alpha}(\bs{m}, \bs{m}'|\bs{s})\|_2^2$ can be estimated as follows:
\begin{align}
     &\rE_{\bs{m}\sim p(\cdot|\bs{s})} \rE_{\bs{m}'\sim p(\cdot|\bs{s})} \|\mathcal{G}^{\alpha}(\bs{m}, \bs{m}'|\bs{s})\|_2^2\nonumber\\
     =& \sum_{j\in \mathcal{C}}\rE_{\bs{m}\sim p(\cdot|\bs{s})} \rE_{\bs{m}'\sim p(\cdot|\bs{s})} \left(\mathcal{G}_j^{\alpha}(\bs{m}, \bs{m}'|\bs{s})\right)^2\nonumber \\
     \leq& V_{\max}(\bs{s}) \sum_{j\in \mathcal{C}}  \bs{s}_j^{2\alpha}(1-\bs{s}_j)^{(2\alpha-1)}+ \bs{s}_j^{2\alpha-1}(1-\bs{s}_j)^{2\alpha}.\label{eqn:proof-8}
\end{align}
Thus, when $\alpha \in [\frac{1}{2},1)$, we have 
\begin{align}
     &\rE_{\bs{m}\sim p(\cdot|\bs{s})} \rE_{\bs{m}'\sim p(\cdot|\bs{s})} \|\mathcal{G}^{\alpha}(\bs{m}, \bs{m}'|\bs{s})\|_2^2\nonumber \\
     \leq& V_{\max}(\bs{s}) \sum_{j\in \mathcal{C}}  \bs{s}_j^{2\alpha}(1-\bs{s}_j)^{(2\alpha-1)}+ \bs{s}_j^{2\alpha-1}(1-\bs{s}_j)^{2\alpha}\nonumber\\  
     \leq&  |\mathcal{C}|V_{\max}(\bs{s}) .\nonumber
\end{align}
The last inequality holds since the term $\bs{s}_j^{2\alpha}(1-\bs{s}_j)^{(2\alpha-1)}+ \bs{s}_j^{2\alpha-1}(1-\bs{s}_j)^{2\alpha}$ is  monotonically decreasing w.r.t. $\alpha\in [\frac{1}{2}, 1)$. 

Therefore, from Property \ref{property:sampling} and \ref{property:flip}, we can see that the variance is bounded for any $\bs{s}$. 

\end{proof}

\begin{remark}
Eqn. (\ref{eqn:proof-2}) and (\ref{eqn:proof-3}) indicate that $H^{\alpha}(\bs{\s})$ is introduced to reduce the variance of the stochastic PGE term $\nabla_{\boldsymbol{s}} \ln{p(\boldsymbol{m}|\boldsymbol{s})}$. Without  $H^{\alpha}(\bs{\s})$ (i.e., $\alpha=0$), from Eqn.(\ref{eqn:proof-8}), we can see that the total variance bound would be 
\begin{align}
      V_{\max}(\bs{s})\sum_{j\in \mathcal{C}}  \frac{1}{(1-\bs{s}_j)}+\frac{1}{ \bs{s}_j}.\nonumber
\end{align}
Because of the sparsity constraints, lots of $\bs{s}_j$ would be close to $0$. Hence, the total variance in this case could be very large. 
\end{remark}

\begin{remark}
Our preconditioning matrix $H^{\alpha}(\bs{s})$ plays a role as adaptive step size. The hyper-parameter $\alpha$ can be used to tune its effect on variance reduction. For a large variance $\nabla_{\boldsymbol{s}} \ln{p(\boldsymbol{m}|\boldsymbol{s})}$ we can use a large $\alpha$. In our experiments, we find that simply letting $\alpha = \frac{1}{2}$ works well. 
\end{remark}
\subsection{Convergence of Our Method}
For the weight update, the convergence can be guaranteed since we use the standard stochastic gradient descent with the gradient calculated via backward propagation. 

For the parameter $\bs{s}$, as stated in Section \ref{sec:VR-PGE}, we update it as:
\begin{align}
\boldsymbol{s}\gets \boldsymbol{s} - \eta \left(\mathcal{L}\left(\boldsymbol{m}\right)- \mathcal{L}\left(\boldsymbol{m}'\right)\right)H^{\alpha}(\boldsymbol{s}) \nabla_{\boldsymbol{s}} \ln{p(\boldsymbol{m}|\boldsymbol{s})}.\label{eqn:update-VR-PGE-appendix}
\end{align}
Let $\Delta \bs{s}(\bs{m}, \bs{m}'|\bs{s}) $ be $\left(\mathcal{L}\left(\boldsymbol{m}\right)- \mathcal{L}\left(\boldsymbol{m}'\right)\right)H^{\alpha}(\boldsymbol{s}) \nabla_{\boldsymbol{s}} \ln{p(\boldsymbol{m}|\boldsymbol{s})}$, we can  have 
\begin{align}
    &\rE_{\bs{m}\sim p(\cdot|\bs{s})} \rE_{\bs{m}'\sim p(\cdot|\bs{s})}\Delta \bs{s}(\bs{m}, \bs{m}'|\bs{s}) \nonumber \\
    =&\rE_{\bs{m}\sim p(\cdot|\bs{s})} \rE_{\bs{m}'\sim p(\cdot|\bs{s})} \left(\mathcal{L}\left(\boldsymbol{m}\right)- \mathcal{L}\left(\boldsymbol{m}'\right)\right)H^{\alpha}(\boldsymbol{s}) \nabla_{\boldsymbol{s}} \ln{p(\boldsymbol{m}|\boldsymbol{s})}\nonumber \\
    =& \rE_{\bs{m}\sim p(\cdot|\bs{s})}  \mathcal{L}\left(\boldsymbol{m}\right)H^{\alpha}(\boldsymbol{s}) \nabla_{\boldsymbol{s}} \ln{p(\boldsymbol{m}|\boldsymbol{s})}- \rE_{\bs{m}\sim p(\cdot|\bs{s})} \rE_{\bs{m}'\sim p(\cdot|\bs{s})}  \mathcal{L}\left(\boldsymbol{m}'\right)H^{\alpha}(\boldsymbol{s}) \nabla_{\boldsymbol{s}} \ln{p(\boldsymbol{m}|\boldsymbol{s})}\nonumber \\
    =&H^{\alpha}(\boldsymbol{s})\rE_{\bs{m}\sim p(\cdot|\bs{s})}  \mathcal{L}\left(\boldsymbol{m}\right) \nabla_{\boldsymbol{s}} \ln{p(\boldsymbol{m}|\boldsymbol{s})}- H^{\alpha}(\boldsymbol{s}) \rE_{\bs{m}'\sim p(\cdot|\bs{s})} \mathcal{L}\left(\boldsymbol{m}'\right)\underbrace{\rE_{\bs{m}\sim p(\cdot|\bs{s})}  \nabla_{\boldsymbol{s}} \ln{p(\boldsymbol{m}|\boldsymbol{s})}}_{I}\nonumber\\
    =& H^{\alpha}(\boldsymbol{s})\rE_{\bs{m}\sim p(\cdot|\bs{s})}  \mathcal{L}\left(\boldsymbol{m}\right) \nabla_{\boldsymbol{s}} \ln{p(\boldsymbol{m}|\boldsymbol{s})}\label{eqn:proof-I} \\
    =& H^{\alpha}(\boldsymbol{s})\nabla_{\boldsymbol{s}}\rE_{\bs{m}\sim p(\cdot|\bs{s})}  \mathcal{L}\left(\boldsymbol{m}\right),\nonumber
\end{align}
where Eqn.(\ref{eqn:proof-I}) holds since term $I= \nabla_{\boldsymbol{s}}\rE_{\bs{m}\sim p(\cdot|\bs{s})} \bs{1} \equiv 0$.

Therefore, we can see that $\Delta \bs{s}(\bs{m}, \bs{m}'|\bs{s})$ is an unbiased gradient estimator associated with an adaptive step size, i.e., our \ref{eqn:VR-PGE} is a standard preconditioned stochastic gradient descent method. Thus, the convergence can be guaranteed.

\subsection{Experiments Verfiying Properties 1 and 2 in \ref{sec:property}}
\begin{figure}[t!]
\begin{center}
\centering  
\includegraphics[width=0.5\linewidth]{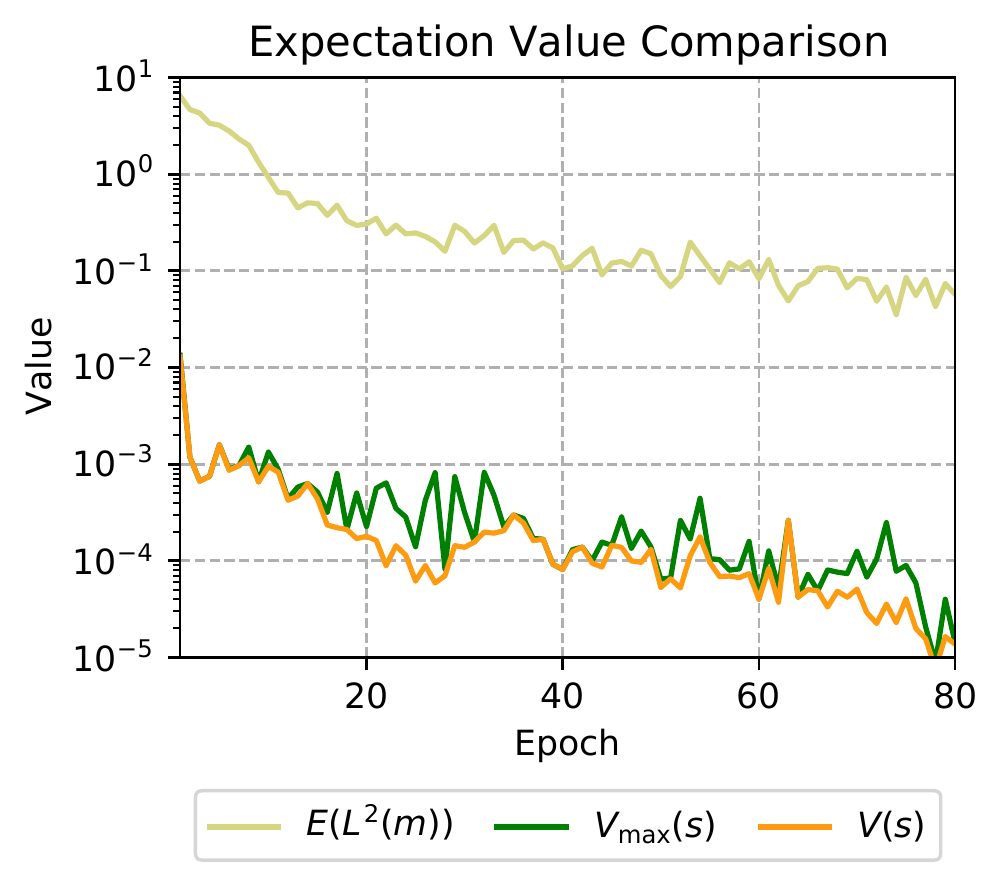}
\end{center}
\caption{Experiments on ResNet32 on CIFAR-10. $V(\bs{s})$ and $V_{\max}(\bs{s})$ are very close during the whole training process and they are smaller than $\rE_{\bs{m}\sim p(\cdot|\bs{s})} \mathcal{L}^2(\bs{m})$ by four orders of magnitude.}\label{fig:appendix-expectation}
\vspace*{-10pt}
\end{figure}

Figure \ref{fig:appendix-expectation} presents the values of $\rE_{\bs{m}\sim p(\cdot|\bs{s})} \mathcal{L}^2(\bs{m})$, $V(\bs{s})$ and $V_{\max}(\bs{s})$ during the training process of ResNet-32 on CIFAR-10. We can see that $V(\bs{s})$ and $V_{\max}(\bs{s})$ are very close during the whole training process and they are smaller than $\rE_{\bs{m}\sim p(\cdot|\bs{s})} \mathcal{L}^2(\bs{m})$ by four orders of magnitude. This verifies our Property \ref{property:sampling} and \ref{property:flip}. 

\section{Experimental Configurations}\label{sec:appendix-experiment}

\textbf{[CIFAR-10/100 Experiments]}
GPUs: 1 for VGG and ResNet and 2 for WideResNet.
Batch Size: 256.
Weight Optimizer: SGD.
Weight Learning Rate: 0.1.
Weight Momentum: 0.9.
Probability Optimizer: Adam.
Probability Learning Rate: \textbf{12e-3}.
WarmUp: \xmark.
Label Smoothing: \xmark.

\textbf{[ImageNet-1K Experiments]}
GPUs: 4.
Batch Size: 256.
Weight Optimizer: SGD.
Weight Learning Rate: 0.256.
Weight Momentum: 0.875.
Probability Optimizer: Adam.
Probability Learning Rate: \textbf{12e-3}.
WarmUp: \cmark.
Label Smoothing: 0.1.

\begin{remark}
The bold-face probability learning rate 12e-3 is the \textbf{only} hyperparameter obtained by grid search on CIFAR-10 experiments and applied directly to larger datasets and networks. Other hyperparameters are applied following the same practice of previous works \cite{ramanujan2020s, kusupati2020soft, liu2018rethinking, zhu2017prune}. The channels of ResNet32 for CIFAR experiments are doubled following the same practice of \cite{wang2020picking}.
\end{remark}

\begin{table*}[htb!]
\caption{Forward/backward time of dense/sparse networks and accompanying properties.}
\begin{center}
{\footnotesize
\begin{tabular}{p{1.65cm}<{\centering} p{1.65cm}<{\centering}p{1.65cm}<{\centering}p{1.65cm}<{\centering}p{1.65cm}<{\centering}p{3cm}<{\centering}}\toprule
Model& Val Acc(\%) &Params(\%)& Forward(min)& Backward(min) & Train-Computational Time(min) \\ \cmidrule(){1-6}
\multirow{6}{*}{VGG-19} & 93.84 & 100.00 & 6.89 & 14.96 & 21.85 (1.00$\times$)\\
      & 93.46 & 23.71 & 6.41 & 7.63 & 14.04 (1.55$\times$)\\
      & 93.11 & 12.75 & 4.89 & 5.54 & 10.43 (2.09$\times$)\\
      & 92.23 & 6.69 & 3.10 & 3.73 & 6.83 (3.20$\times$)\\
      & 90.82 & 3.06 & 2.15 & 2.71 & 4.86 (4.50$\times$)\\
      & 87.97 & 0.80 & 1.27 & 1.68  & 2.95 (7.41$\times$)\\ 
      \bottomrule
\end{tabular}
}
\end{center}
\label{tab:time}
\end{table*}

\section{Calculation Schemes on Train-cost Savings and Train-computational Time}\label{sec:appendix-calculation}
\subsection{Train-cost Savings}
The train-cost of vanilla dense training can be computed as two parts: in forward propagation, calculating the loss of weights and in backward propagation, calculating the gradient of weights and gradient of the activations of the previous layers. The FLOPs of backward propagation is about 2$\sim$3 times of forward propagation \cite{baydin2018automatic}. In the following calculation, we calculate the FLOPs of forward propagation concretely and consider FLOPs of backward propagation 2 times of forward propagation for simplicity.

\textbf{[GrowEfficient]} The forward propagation of dense network is $f_{D}$. The forward propagation of GrowEfficient is partially sparse with FLOPs being $f_{S}$ and backward propagation is dense. Therefore the train-cost saving is computed as $\frac{f_D+2f_D}{f_S+f_D}=\frac{3}{2+f_S/f_D}$, upper-bounded by $\frac{3}{2}$.

\textbf{[Ours]} The forward propagation of dense network is $f_{D}$. The forward propagation and backward propagation is totally sparse. The FLOPs of forward propagation is $f_S$ and the FLOPs of backward propagation is $2*f_S$. The forward propagation has to be computed two times. Therefore the train-cost saving is computed as $\frac{f_D+2*f_D}{2*f_S+2*f_S}=\frac{3}{4f_S/f_D}$. Actually, $f_S/f_D$ is roughly equal to $\rho^2$, leading to drastically higher train-cost savings.

\subsection{Train-computational Time}
The calculation of train-computational time focuses on the forward and backward propagation of dense/sparse networks. For both of the dense and sparse networks, we sum up the computation time of all the forward and backward propagation in the training process as the train-computational time. The detailed time cost  is presented in Table \ref{tab:time}. We can see that we can achieve significant speedups in computational time. 

\section{Potentials and Limitations of This Work}\label{sec:appendix-discuss}
\textbf{[On Computational Cost Saving]} Although our method needs two forward propagation in each iteration, we have to point out that  our method can achieve significant computational cost saving. The reason is that our forward and backward is completely sparse, whose computational complexity is roughly $\rho^2 *100\%$ of the conventional training algorithms with $\rho$ being the remain ratio of the channels.

\textbf{[On Exploring Larger Networks]} About the potential of our method in exploring larger networks, we'd like to clarify the following three things:
\begin{itemize}
    \item[1. ]The memory cost of the structure  parameters $\bs{s}$ is negligible compared with the original weight $\bs{w}$ as each filter is associated with only one structure parameter, therefore our $\bs{s}$ would hardly increase the total memory usage.
    \item[2. ]Although in our method, we need to store the parameter of the full model, this would not hinder us from exploring larger networks. The reason is that, in each iteration, we essentially perform forward and backward propagation on the sparse subnetwork. More importantly, we find that reducing the frequency of sampling subnetwork, e.g., sample a new subnetwork for every 50 iterations, during training would not affect the final accuracy. In this way, we can store the parameters of the full model on CPU memory and store the current subnetwork on GPU, and synchronize the parameters' updates to  the full model  only when we need to resample a new subnetwork. Hence, our method has great potentials in exploring larger deep neural networks. We left such  engineering implements as the future work and we also welcome the engineers in the community to implement our method more efficiently. 
    \item[3. ]In exploring larger networks, the channel remain ratio $\rho$ can be much smaller than the one in the experiments in the main text. Notice that our method can reduce the computational complexity to $\rho^2*100\%$ of the full network.  It implies that, in this scenario, the potential of our method can be further stimulated. We left this evaluation as future work after more efficient implementation as discussed above. 
\end{itemize}


\clearpage
\newpage
{
	\small
\bibliographystyle{plainnat}
	\bibliography{egbib}
}

\end{document}